\documentclass[10pt,twocolumn,letterpaper]{article}

\usepackage{iccv}
\usepackage{times}
\usepackage{epsfig}
\usepackage{graphicx}
\usepackage{amsmath}
\usepackage{amssymb}
\usepackage{mathtools}
\usepackage{gensymb}
\usepackage[dvipsnames]{xcolor}
\usepackage{color, soul}

\usepackage{tabularx, booktabs}
\usepackage{subcaption}

\usepackage[breaklinks=true,bookmarks=false]{hyperref}

\iccvfinalcopy 

\def\iccvPaperID{3104} 

\ificcvfinal\fi

\begin{document}

\title{Key.Net: Keypoint Detection by Handcrafted and Learned CNN Filters}


\author{Axel Barroso-Laguna$^{1}$\\
\and
Edgar Riba$^{2, 3}$\\
\and
Daniel Ponsa$^{2}$\\
\and
Krystian Mikolajczyk$^{1}$\\
\and
$^{1}$Imperial College London \hspace{0.8cm} $^{2}$Computer Vision Center \hspace{0.8cm} $^{3}$Arraiy, Inc. \\
{\{axel.barroso17, k.mikolajczyk\}@imperial.ac.uk} \hspace{0.5cm} {\{eriba, daniel\}@cvc.uab.es}
\and
}

\maketitle
\ificcvfinal\thispagestyle{empty}\fi

\begin{abstract}

 We introduce a novel approach for keypoint detection task that combines handcrafted and learned CNN filters within a shallow multi-scale architecture. Handcrafted filters provide anchor structures for learned filters, which localize, score and rank repeatable features. Scale-space representation is used within the network to extract keypoints at different levels. We design a loss function to detect robust features that exist across a range of scales and to maximize the repeatability score. 
 Our Key.Net model is trained on data synthetically created from ImageNet and evaluated on HPatches benchmark. Results show that our approach outperforms state-of-the-art detectors in terms of repeatability, matching performance and complexity.


\end{abstract}

\section{Introduction}

Research advances in local feature detectors and descriptors led to remarkable improvements in areas such as image matching, object recognition, self-guided navigation or 3D reconstruction. 
Although the general direction of image matching methods is moving towards learned based systems, the advantage of learning methods over handcrafted ones has not been clearly demonstrated in keypoint detection \cite{Karel_Vedaldi_BMVC_18}. 
In particular, Convolutional Neural Networks (CNNs) were able to significantly reduce matching error in local descriptors~\cite{HPatches}, despite the impractical inefficiency of the initial techniques~\cite{MatchNet15,Zagoruyko15}. 
These works stimulated further research efforts and resulted in improved efficiency of CNN based descriptors, 
on the contrary, on top of the limited success of learned detectors, a general trend towards dense rather than sparse representation and matching put aside local feature detectors. 
However, the growing popularity of augmented reality (AR) headsets, as well as AR smartphone apps, has drawn more attention to reliable and efficient local feature detectors that could be used for surface estimation, sparse 3D reconstruction, 3D model acquisition or objects alignment, among others. \par
\begin{figure}
 \centering
 \hspace*{-1cm}    
   \includegraphics[scale=0.55]{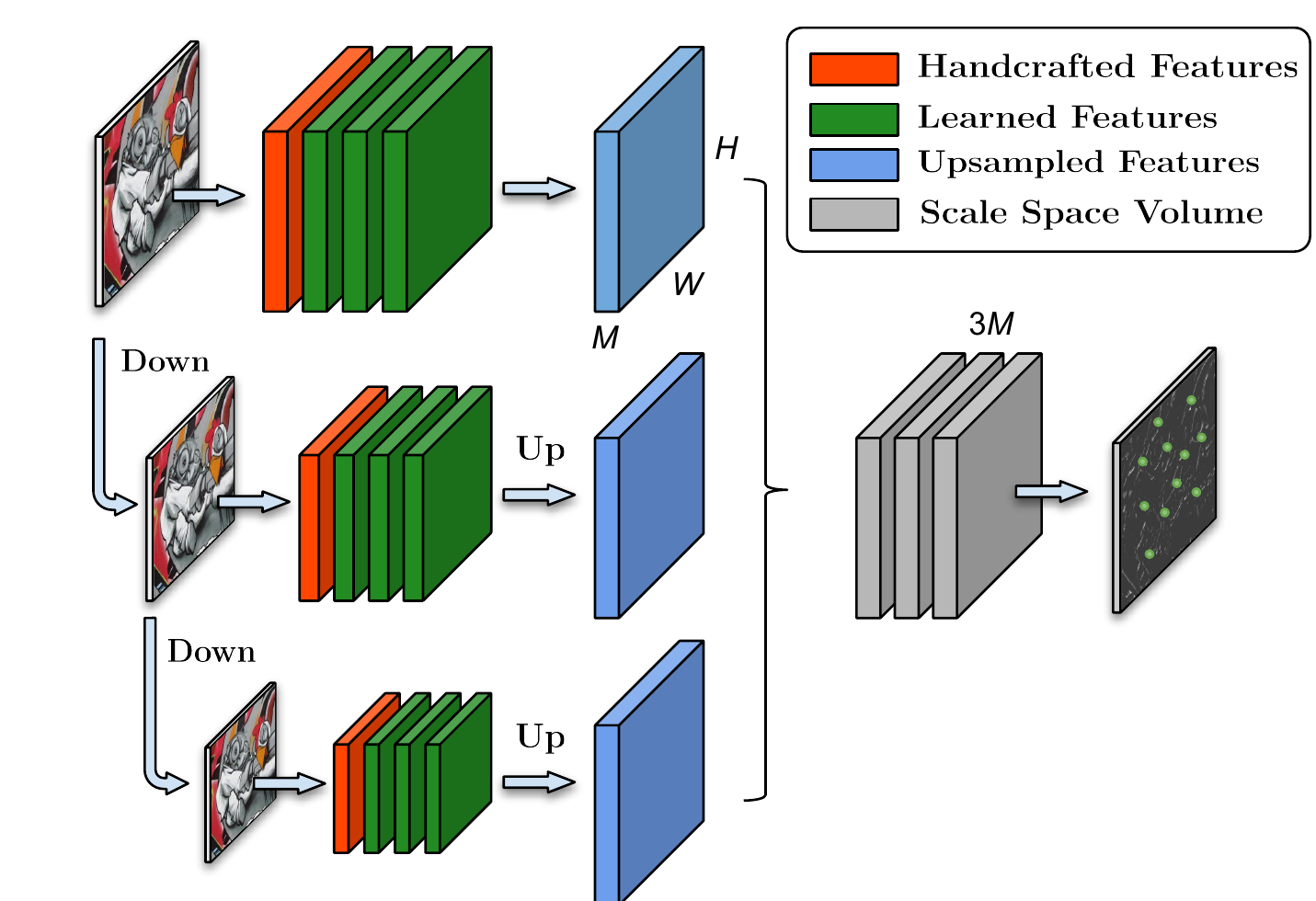}
   \vspace{-0.20cm}
    \caption{The proposed Key.Net architecture combines handcrafted and learned filters to extract features at different scale levels. Feature maps are upsampled and concatenated. Last learned filter combines the Scale Space Volume to obtain the final response map.}
    \label{fig:net_architecture}
\end{figure} 

Traditionally, local feature detectors were based on engineered filters. For instance, approaches such as Difference of Gaussians~\cite{DoG}, Harris-Laplace or Hessian-Affine~\cite{mikolajczykIJCV2004} use combinations of image derivatives to compute feature maps, which is remarkably similar to the operations in trained CNN's layers. Intuitively, with just a few layers, a network could mimic the behavior of traditional detectors by learning the appropriate values in its convolutional filters. However, unlike the success with CNNs based local image descriptors, the improvements upon handcrafted detectors offered by recently proposed fully CNN based methods~\cite{LIFT,DeTone_MagicPoint17,Karel_Vedaldi_ECCV_16, Zhang_Felix_CVPR_17,OnoSerra18} are limited in terms of widely accepted metrics such as repeatability. One of the reasons is their low accuracy when estimating the affine parameters of the feature regions. Robustness to scale variations seems particularly problematic while other parameters such as dominant orientation can be regressed well by CNNs~\cite{Yi_Verdie_Fua_Lepetit_CVPR16, LIFT}. 
This motivates our novel architecture, termed Key.Net, that makes use of handcrafted and learned filters as well as a multi-scale representation. 
The Key.Net architecture is illustrated in figure~\ref{fig:net_architecture}. 
Introducing handcrafted filters, which act as soft anchors, makes possible to reduce the number of parameters used by state-of-the-art detectors while maintaining the performance in terms of repeatability. The model operates on multi-scale representation of full-size images and returns a response map containing the keypoint score for every pixel. The multi-scale input allows the network to propose stable keypoints across scales thus providing robustness to scale changes. 

Ideally, a robust detector is able to propose the same features for images that undergo different geometric or photometric transformations. A number of related works have focused their objective function to address this issue, although they were based either on local patches~\cite{Karel_Vedaldi_ECCV_16, Zhang_Felix_CVPR_17} or global map regression loss~\cite{detone2017superpoint, TILDE, OnoSerra18}. In contrast, we extend the covariant constraint loss to a new objective function that combines local and global information. We design a fully differentiable operator, Multi-scale Index Proposal, that proposes keypoints at multi-scale regions. We extensively evaluate the method in recently introduced HPatches benchmark~\cite{HPatches} in terms of accuracy and repeatability according to the protocol from~\cite{mikolajczykpami2005}.

In summary, our contributions are the following: a) a keypoint detector that combines handcrafted and learned CNN features, b) a novel multi-scale loss and operator for detecting and ranking stable keypoints across scales, c) a multi-scale feature detection with shallow architecture.

The rest of the paper is organized as follows. We review the related work in section~\ref{sec:Relatedwork}. Section~\ref{sec:architecture} presents our proposed hybrid Key.Net architecture of handcrafted and learned CNNs filters and section~\ref{sec:loss_functions} introduces the loss. Implementation and experimental details are given in section~\ref{sec:experimental_evaluation} and the results are presented in section~\ref{sec:results}.

\begin{figure*}[!htb]
    \centering
    \includegraphics[scale=0.5]{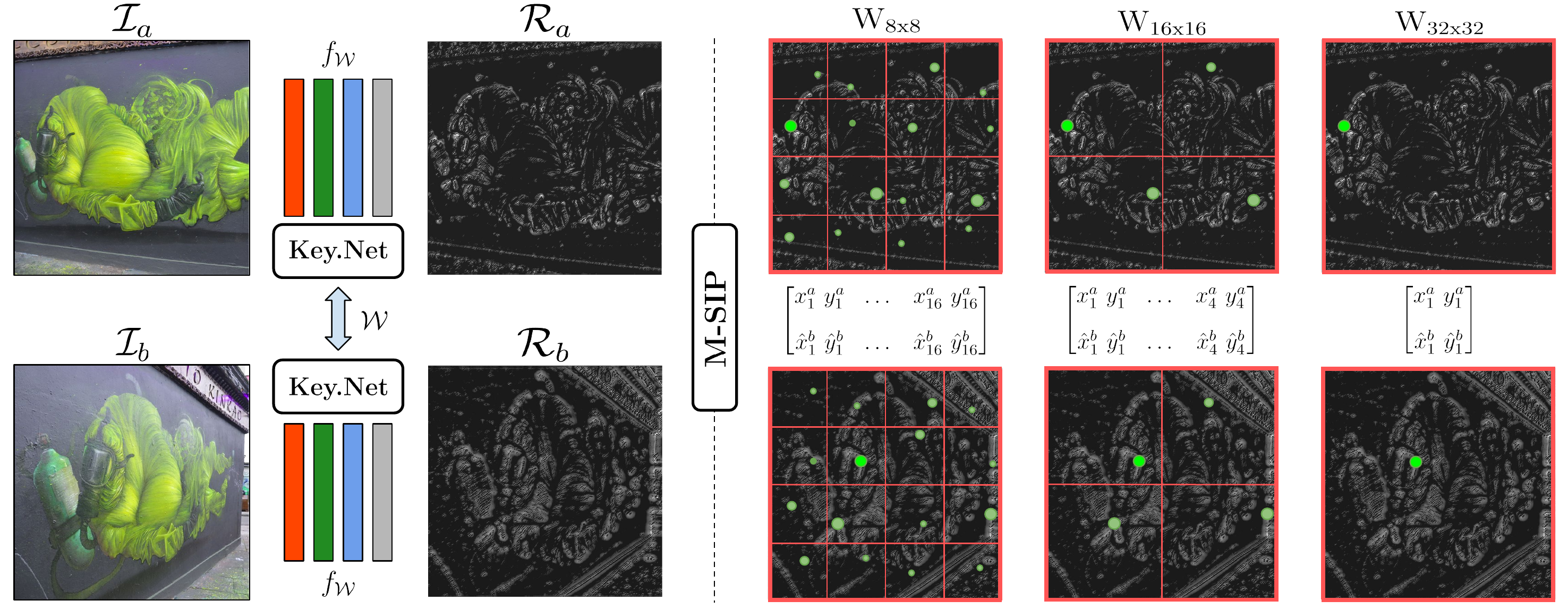}
    \vspace{-0.20cm}
    \caption{Siamese training process. Image $I_a$ and $I_b$ go through Key.Net to generate their response maps, $R_a$ and $R_b$. M-SIP proposes interest point coordinates for each one of the windows at multi-scale regions. The final loss function is computed as a regression of coordinate indexes from $I_a$ and local maximum coordinates from $I_b$. Better visualize in color.}
    
    \label{fig:trainingfigure}
\end{figure*}

\section{Related Work}
\label{sec:Relatedwork}
There are many surveys that extensively discuss feature detection methods~\cite{Karel_Vedaldi_BMVC_18, TuytelaarsMikolajczyk2007}. 
We present related works in two main categories: handcrafted and learned based. 


\subsection{Handcrafted Detectors}
Traditional feature detectors localize geometric structures through engineered algorithms, which are often referred to as handcrafted. Harris \cite{Harris} and Hessian \cite{Hessian} detectors used first and second order image derivatives to find corners or blobs in images. Those detectors were further extended to handle multi-scale and affine transformations \cite{mikolajczykIJCV2004, comparasionofdetector}. Later, SURF \cite{SURF} accelerated the detection process by using integral images and an approximation of the Hessian matrix. Multi-scale improvements were proposed in KAZE \cite{KAZE} and its extension, A-KAZE \cite{AKAZE}, where Hessian detector was applied to a non-linear diffusion scale space in contrast to widely used Gaussian pyramid. Although corner detectors proved to be robust and efficient, other methods seek alternative structures within images. SIFT \cite{DoG} looked for blobs over multiple scale levels, 
and MSER \cite{MSER} segmented and selected stable regions as keypoints.


\subsection{Learned Detectors}
The success of learned methods in general object detection and feature descriptors motivated the research community to explore similar techniques for feature detectors. 
FAST \cite{FAST} was one of the first attempts to use machine learning to derive a corner keypoint detector. Further works extended FAST by optimizing it \cite{FASTER}, adding a descriptor \cite{BRISK} or orientation estimation \cite{ORB}. 

Latest advances in CNNs also made an impact on feature detection. TILDE \cite{TILDE} trained multiple piece-wise linear regression models to identify interest points that are robust under severe weather and illumination changes. \cite{Karel_Vedaldi_ECCV_16} introduced a new formulation to train a CNN based on feature covariant constraints. 
Previous detector was extended in \cite{Zhang_Felix_CVPR_17} by adding predefined detector anchors, showing improved stability in training. \cite{DeTone_MagicPoint17} presented two networks, MagicPoint, and MagicWarp, which first extracted salient points and then a parameterized transformation between pairs of images. MagicPoint was extended in \cite{detone2017superpoint} to SuperPoint, which included a salient detector and descriptor.
LIFT \cite{LIFT} implemented an end-to-end feature detection and description pipeline, including the orientation estimation for every feature. Quadruple image patches and a ranking scheme of point responses as cost function were used in \cite{savinov2016quad} to train a neural network. In \cite{Georgakis_Karanam_CVPR18}, authors proposed a pipeline to automatically sample positive and negative pairs of patches from a region proposal network to optimize jointly point detections and their representations. Recently, LF-Net \cite{OnoSerra18} estimated position, scale and orientation of features by optimizing jointly the detector and descriptor. \par

In addition to the above presented learned detectors, CNN architectures also were deployed to optimize the matching stage. \cite{Hartmann_Havlena_CVPR14} learned to predict which features and descriptors were matchable. More recently, \cite{Yi_Trulls_Good_Corr_CVPR18} introduced a network to learn to find good correspondences for wide-baseline stereo. Furthermore, other CNNs also studied to perform tasks beyond detection or matching. In \cite{Yi_Verdie_Fua_Lepetit_CVPR16}, the architecture assigned orientations to interest points and AffNet \cite{Mishkin_Radenovic_Matas_AffNet_18} used the descriptor loss to learn to predict the affine parameters of a local feature. 

\label{sec:Method}



\section{Key.Net Architecture}
\label{sec:architecture}
Key.Net architecture combines successful ideas from handcrafted and learned methods namely gradient-based feature extraction, learned combinations of low-level features and multi-scale pyramid representation. 


\subsection{Handcrafted and Learned Filters}
\label{sec:handcrafted_filters}
The design of the handcrafted filters is inspired by the success of Harris \cite{Harris} and Hessian \cite{Hessian} detectors, which used first and second order derivatives to compute the salient corner responses. A complete set of derivatives is called {\em LocalJet}~\cite{Koendering-FlorackLocalJet} and they approximate the signal in the local neighborhood as known from Taylor expansion:
\begin{equation}
I_{i_1,\ldots,i_n}=I_0\ast\partial_{i_1,\ldots,i_n}g_{\sigma}(\vec{x}),
\label{eq:Taylor}
\end{equation}
where $g_{\sigma}$ denotes the Gaussian of width $\sigma$ centered at $\vec{x}=\vec{0}$, and $i_n$ denotes the direction. Higher order derivatives i.e., $n>2$ are sensitive to noise and require large kernels, we, therefore, include derivatives and their combinations up to the second order only: 

\begin{itemize}
	\item \textbf{First Order}. From image $I$ we derive $1$st order gradients $I_x$ and $I_y$. In addition, we compute $I_x*I_y$, ${I_x}^2$ and ${I_y}^2$ as in the second moment matrix of Harris detector~\cite{Harris}.
    \item \textbf{Second Order}. From image $I$, $2$nd order derivatives $I_{xx}$, $I_{yy}$ and $I_{xy}$ are also included as in the Hessian matrix used in Hessian and DoG detectors \cite{HarrisLaplace,DoG}.  Since Hessian detector uses the determinant of the Hessian matrix we add  $I_{xx}*I_{yy}$ and $I_{xy}^2$. 
     \item \textbf{Learned.}  A convolutional layer with $M$ filters, a batch normalization layer and a ReLU activation function form a learned block.

\end{itemize}
The hardcoded filters reduce the number of total learnable parameters to train the architecture, improving the stability and convergence during backpropagation.
%



\subsection{Multi-scale Pyramid }
\label{sec:pyramid}


We design our architecture to be robust to small scale changes without the need for computing several forward passes. 
As illustrated in figure~\ref{fig:net_architecture}, the network includes three scale levels of the input image which is blurred and downsampled by a factor of $1.2$. All the feature maps resulting from the handcrafted filters are concatenated to feed the stack of learned filters in each of the scale levels. All three streams share the weights, such that the same type of anchors result from different levels and form the set of candidates for final keypoints. 
Feature maps from all scale levels are then upsampled, concatenated and fed to the last convolutional filter to obtain the final response map. 


\section{Loss Functions}
\label{sec:loss_functions}

In supervised training, the loss function relies on the ground truth. In the case of keypoints, ground truth is not well defined as keypoint locations are useful as long as they can be accurately detected regardless of geometric or photometric image transformation. 
Some learned detectors \cite{Karel_Vedaldi_ECCV_16,savinov2016quad,OnoSerra18} train the network to identify keypoints without constraining their locations, where only the homography transformation between images is used as ground truth to calculate the loss as a function of keypoints repeatability. 

Other works~\cite{TILDE,detone2017superpoint,Zhang_Felix_CVPR_17} show the benefits of using anchors to guide their training. Although anchors make the training more stable and lead to better results, they prevent the network from proposing new keypoints in case there is no anchor in the proximity. 

In contrast, the handcrafted filters in Key.Net provide a weak constraint with the benefit of the anchor-based methods while allowing the detector to propose new stable keypoints. In our approach, only the geometric transformation between images is required to guide the loss. \par


\subsection{Index Proposal Layer}
\label{sec:single_index_proposal_layer}


This section introduces the Index Proposal (IP) layer, which is extended to its multi-scale version in section \ref{sec:multi_index_proposal_layer}. \par


Extracting coordinates for training keypoint detectors has been widely studied and showed great improvements: \cite{LIFT, Karel_Vedaldi_ECCV_16, Zhang_Felix_CVPR_17} extracted coordinates in a patch level, SuperPoint \cite{detone2017superpoint} used a channel-wise softmax to get maxima belonging to fix grids of $8x8$, and \cite{keypointnet_nips_18} used a spatial softmax layer to compute the global maxima of a feature map, obtaining one keypoint candidate per feature map. In contrast to previous methods, the IP layer is able to return multiple global keypoint coordinates centered on local maxima from a single image without constraining the number of keypoints to the depth of the feature map \cite{keypointnet_nips_18} or the size of the grid \cite{detone2017superpoint}.

 Similarly to handcrafted techniques, keypoint locations are indicated by local maxima of the filter response map $\mathcal{R}$ output by Key.Net. Spatial softmax operator is an effective method for extracting the location of a soft maximum within a window \cite{LIFT, keypointnet_nips_18, OnoSerra18, detone2017superpoint}. Therefore, to ensure that the IP layer is fully differentiable, we rely on spatial softmax operator to obtain the coordinates of a single keypoint per window. Consider a window $w_i$ of size $N \times N$ in $\mathcal{R}$,
with the score value at each coordinate $[u,v]$ within the window, exponentially scaled and normalized: 
\begin{equation}
m_{i}(u,v) = \dfrac{e^{w_i(u, v)}}{\sum_{\substack{j, k}}^{N} e^{w_i(j, k)}}.
\label{eq:spatial_softmax}
\end{equation}
Due to exponential scaling the maximum dominates and the expected location calculated as the weighted average $[\Bar{u_i},\Bar{v_i}]$ gives an approximation of the maximum coordinates:
\begin{equation}
[x_i,y_i]^{T}=[\Bar{u_i}, \Bar{v_i}]^{T} =\sum_{\substack{u, v}}^{N} [W \odot m_{i}, W^T \odot m_{i}]^{T} + c_{w},
\label{eq:Index_Proposal}
\end{equation}
where $W$ is a kernel of size $N \times N$ with index values $j=1:N$ along its columns, pointwise product $\odot$, and $c_{w}$ is the top-left corner coordinates of window $w_{i}$.
This is similar to non-maxima suppression (NMS) but unlike NMS, the IP layer is differentiable and it is a weighted average of the global maximum of the window rather than the exact location of it.
Depending on the base of the power expression in equation~\ref{eq:spatial_softmax}, multiple local maxima may have a more or less significant effect on the resulting coordinates.  


A detector is covariant if same features are detected under varying image transformations. 
Covariant constraint was formulated as a regression problem in \cite{Karel_Vedaldi_ECCV_16}. 
Given images $I_{a}$ and $I_{b}$, and ground truth homography $H_{b,a}$ between them,  the loss $\mathcal{L}$ is based on the squared difference between points extracted by IP layer and actual maximum coordinates (NMS) in corresponding windows from $I_{a}$ and $I_{b}$ :
\begin{equation*}
\mathcal{L}_{IP}(I_a,I_b,H_{a,b}, N) = \sum_{\substack{i}} \alpha_i\| [{x_i}, {y_i}]^T_{a} - H_{b,a}[{\hat{x}_i}, {\hat{y}_i}]^T_{b}\|^2, \\
\end{equation*}
\begin{equation}
\textrm{and  } \textrm{ } \textrm{ } \alpha_i =  \mathcal{R}_{a}({x_i}, {y_i})_{a} + \mathcal{R}_{b}({\hat{x}_i}, {\hat{y}_i})_{b},
\label{eq:context_losses}
\end{equation}
where $\mathcal{R}_{a}$ and $\mathcal{R}_{b}$ are the response map of $I_a$ and $I_b$ with coordinates related by the homography $H_{b,a}$. We skip homogeneous coordinates for simplicity. Parameter $\alpha_{i}$ controls the contribution of each location based on its score value, thus computing the loss for significant features only. As NMS is non-differentiable, gradients are only back-propagated where IP layer is applied, therefore, we switch $I_{a}$ and $I_{b}$ and combine both losses to enforce consistency. 


\subsection{Multi-scale Index Proposal Layer}
\label{sec:multi_index_proposal_layer}

\begin{figure}[t]
\vspace{-0.10cm}
 \hspace*{-0.4cm} 
 \centering
   \includegraphics[scale=0.36]{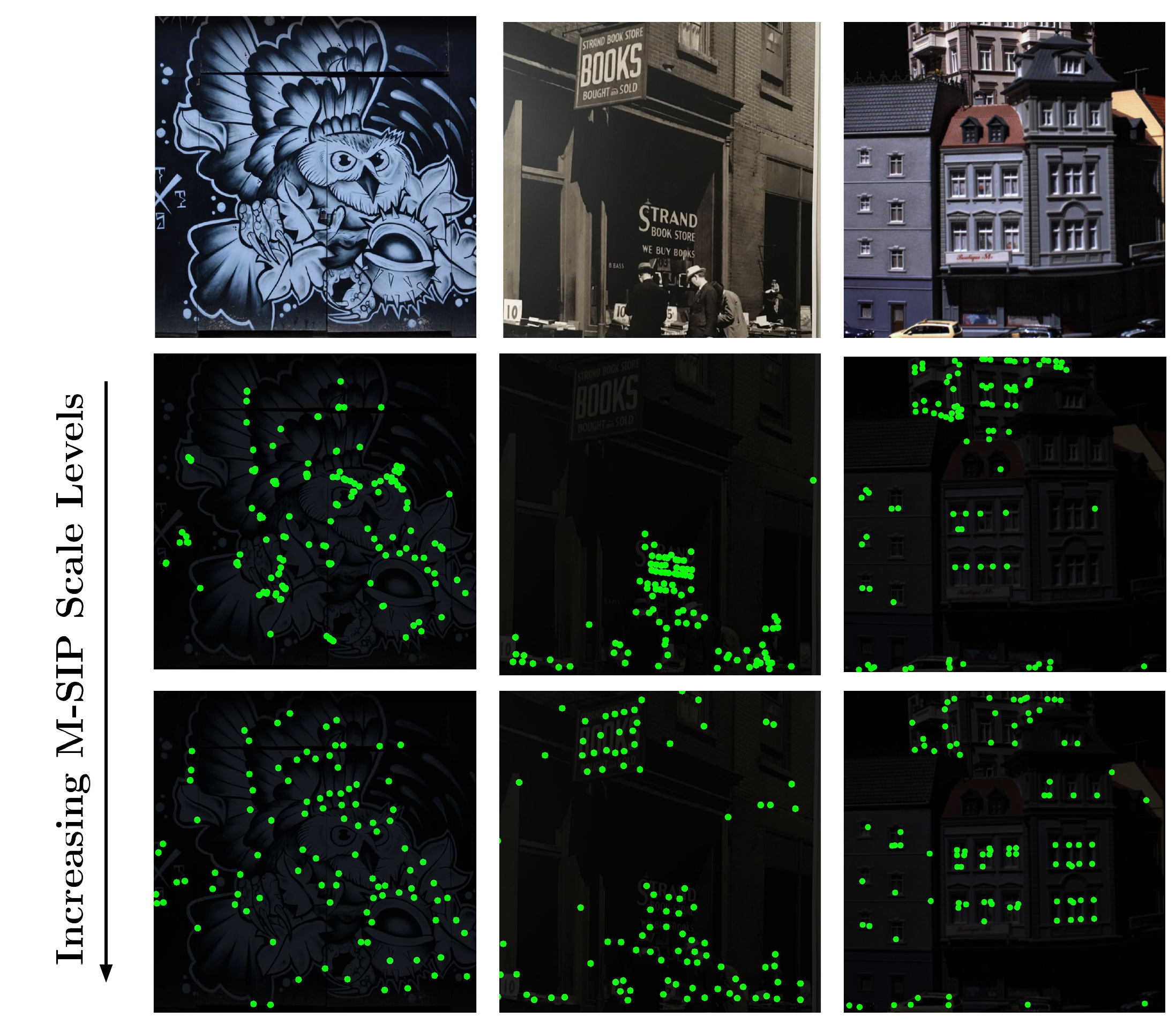}
   \vspace{-0.2cm}
    \caption{Keypoints obtained after adding larger context windows to M-SIP operator. The points that are more stable remain as the M-SIP operator increases its window size. Feature maps in the middle row contain points around edges or non discriminative areas, while bottom row shows detections that are more robust under geometric transformations.}
    \label{fig:M-SIP_Scale_Levels}
\end{figure}

IP layer returns one location per window, therefore, the number of keypoints per image strongly depends on the predefined window size $N$, in particular, with an increasing size only a few dominant keypoints survive in the image.
In \cite{DSPSIFT_soatto}, authors demonstrated improved performance of local features by accumulating image features not only within a spatial window but also within the neighboring scales.
We propose to extend IP layer loss by incorporating multi-scale representation of a local neighborhood. Multiple window sizes encourage the network to find keypoints that exist across a range of scales. The additional benefit of including larger windows is that other keypoints within the window can act as anchors for the estimated location of the dominant keypoint. Similar idea proved successful in~\cite{Local_affine_frames_Matas}, where stable region boundaries are used. 

We, therefore, propose the Multi-Scale Index Proposal (M-SIP) layer. 
M-SIP splits multiple times the response map into grids, each with a window size of $N_s \times N_s$ and computes the candidate keypoint position for each window as shown in figure~\ref{fig:trainingfigure}.  
Our proposed loss function is the average of covariant constraint losses from all scale levels: 
\begin{equation}
\mathcal{L}_{MSIP}(I_{a}, I_{b}, H_{a,b}) =  \sum_{\substack{s}}\lambda_{s} \mathcal{L}_{IP}(I_{a}, I_{b}, H_{a,b},N_s),
  \label{eq:context_losses_m}
\end{equation}
where $s$ is the index of the scale level with $N_s$ as window size, $\mathcal{L}_{IP}$ is the covariant constraint loss and $\lambda_{s}$ is the control parameter at scale level $s$, that decreases proportionally to the increasing window area as larger windows lead to a larger loss, which is somewhat similar to the scale-space normalisation~\cite{mikolajczykIJCV2004}. \par 

The combination of different scales imposes an intrinsic process of simultaneous scoring and ranking of keypoints within the network. In order to minimize the loss, the network must learn to give higher scores to robust features that remain dominant across a range of scales. Figure~\ref{fig:M-SIP_Scale_Levels} shows different response maps for increasing window size. 

\section{Experimental Settings}
\label{sec:experimental_evaluation}

In this section, we present implementation details, metrics and the dataset used for evaluating the method.

\subsection{Training Data}
\label{sec:create_dataset}

\begin{figure}[t]
 \hspace*{-0.1cm} 
 \vspace{-0.4cm}
    \centering
    \includegraphics[scale=0.53]{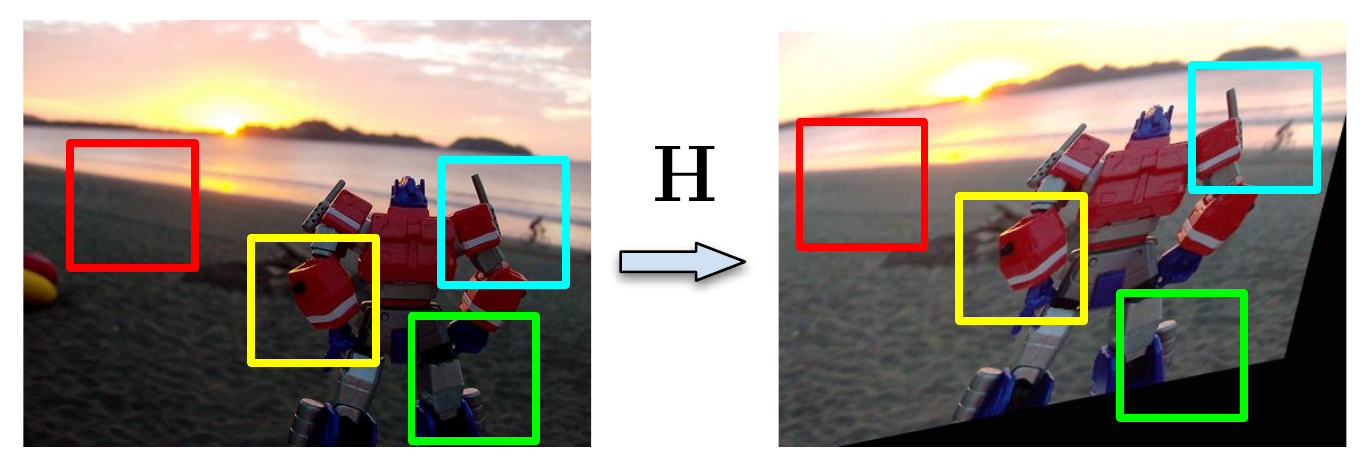}
    \vspace{-0.0cm}
    \caption{We apply random geometric and photometric transformations to images and extract pairs of corresponding regions as the training set. 
    Red crop is discarded by checking the response of the handcrafted filters.
    }
    \label{fig:dataset}
\end{figure}

We generate a synthetic training set from ImageNet ILSVRC 2012 dataset. We apply random geometric transformations to images and extract pairs of corresponding regions as our training set. The process is illustrated in figure~\ref{fig:dataset}. The parameters of the transformations are: scale $[0.5, 3.5]$, skew  $[-0.8, 0.8]$ and rotation $[-60\degree, 60\degree]$. Textureless regions are not discriminative, therefore, we discard them by checking if the response of any of the handcrafted filters is lower than a threshold. We modify the contrast, brightness and hue value in HSV space to one of the images to improve network's robustness against illumination changes. In addition, for each pair, we generate binary masks that indicate the common area between images. Those masks are used in training to avoid regressing indexes of keypoints that are not present in the common region. There are 12,000 image pairs of size 192 $\times$ 192. We use 9,000 of them as the training data and 3,000 as validation set.


\subsection{Evaluation Metrics}
\label{subsec:Evaluation}
We follow the evaluation protocol proposed in \cite{mikolajczykpami2005} and improved in the follow up works~\cite{LIFT,Karel_Vedaldi_ECCV_16,Zhang_Felix_CVPR_17,Karel_Vedaldi_BMVC_18}. 
 Repeatability score for a pair of images is computed as the ratio between the number of corresponding keypoints and the lower number of keypoints detected in one of the two images. We fix the number of extracted keypoints to compare across methods and allow each keypoint to match only once as in \cite{FASTER, TILDE}. In addition, as exposed by \cite{Karel_Vedaldi_BMVC_18}, we address the bias from the magnification factor that was applied to accelerate the computation of the overlap error between multi-scale keypoints.
Keypoints are identified by spatial coordinates and scales at which the features were detected.  To identify corresponding keypoints we compute the Intersection-over-Union error, $\epsilon_{IoU}$, between the areas of the two candidates.
To evaluate the accuracy of keypoint location and scale independently, we perform two sets of experiments. One is based on the detected scales and the other assumes the scales are correctly detected by using the ground truth parameters. 
In our benchmark, we use top 1,000 interest points that belong to the common region between images and a match is considered correct when $\epsilon_{IoU}$ is smaller than 0.4 i.e., the overlap between corresponding regions is more than 60\%. The scales are normalized as in \cite{Karel_Vedaldi_BMVC_18}, which sets the larger size in a pair of points to 30 pixels, and rescales the other one accordingly. Non-maxima suppression of $15 \times 15$ is performed at inference time during evaluation. HPatches~\cite{HPatches} dataset is used for testing. HPatches contains 116 sequences, which are split between viewpoint and illumination transformations, 59 and 57 sequences respectively. HPatches offers predefined image patches for evaluating descriptors, instead, we use full images for evaluating keypoint detectors. 

\begin{figure*}[ht]
\vspace{-0.10cm}
\begin{minipage}[b]{.5\textwidth}
    \centering
    \begin{tabular}{ccccclc}
        \hline
        \multicolumn{5}{c}{M-SIP Region Sizes} & \multicolumn{1}{c}{} & \multicolumn{1}{c}{}\\ 
        \cline{1-5} 
        W{$_{8\text{x}8}$} & W{$_{16\text{x}16}$} & W$_{24\text{x}24}$ & W$_{32\text{x}32}$ & W$_{40\text{x}40}$ & & Repeatability \\
        \hline
        \checkmark & - & - & - & - && 70.5  \\
        \checkmark & \checkmark & - & - & - && 74.6  \\
        \checkmark & \checkmark & \checkmark & - & - && 76.8  \\
        \checkmark & \checkmark & \checkmark & \checkmark & - && 77.6  \\
        - & - & - & - & \checkmark && 65.7 \\
        - & - & - & \checkmark & \checkmark && 71.4 \\
        - & - & \checkmark & \checkmark & \checkmark && 73.2  \\
        - & \checkmark & \checkmark & \checkmark & \checkmark && 74.9  \\
        \checkmark & \checkmark & \checkmark & \checkmark & \checkmark && \textbf{79.1}  \\
    \end{tabular}
\end{minipage}
\begin{minipage}[b]{.57\textwidth}
    \centering
    \includegraphics[scale=0.4]{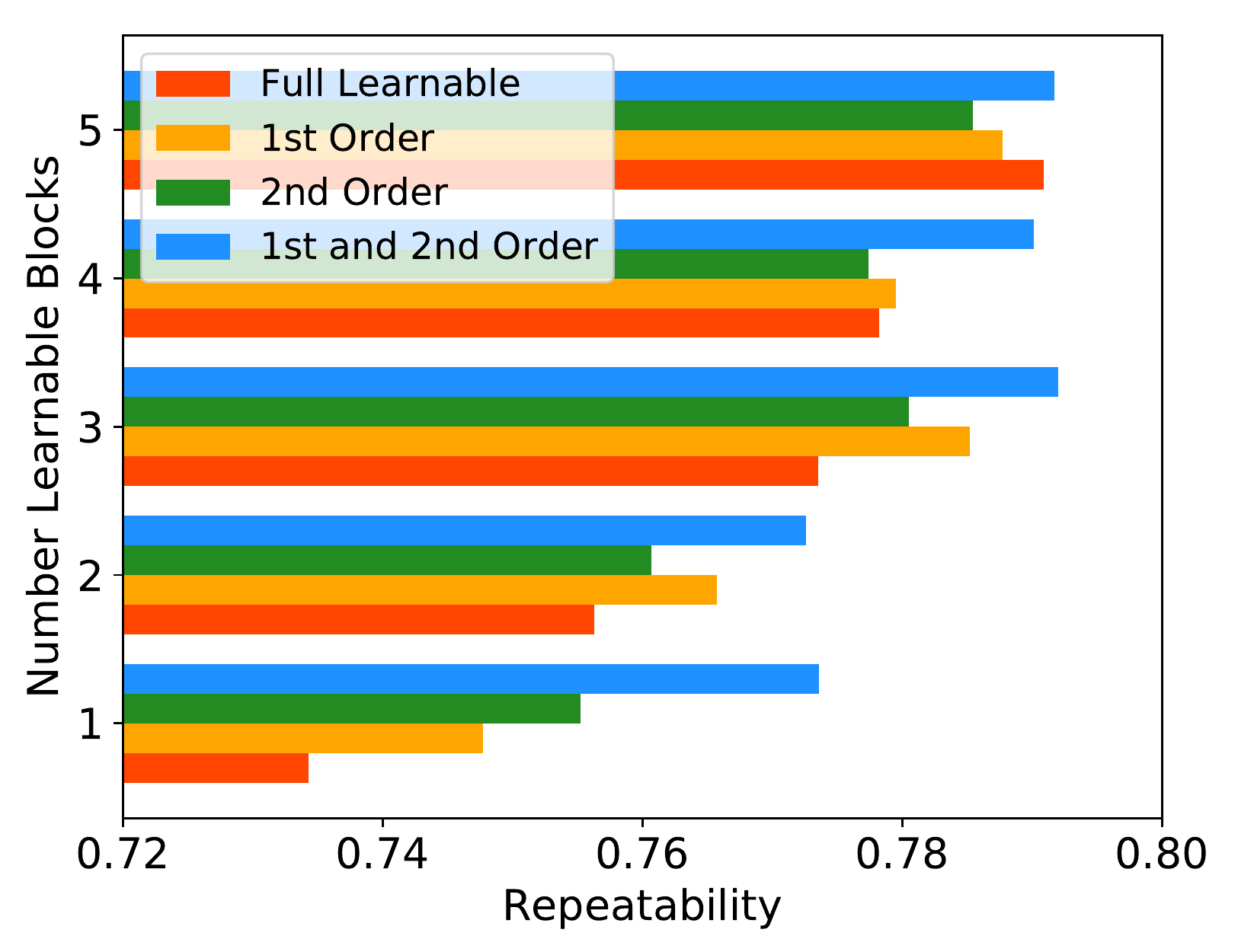}
    \vspace{-2.45cm}
\end{minipage}
\vspace{-0.25cm}
\caption{\textbf{Left}: Comparison of repeatability results for several levels in the M-SIP operator. We show different combinations of context losses as the final loss,  from smaller to larger regions. The best result is obtained when using five window sizes from $8\times 8$ up to $40\times 40$. \textbf{Right}: Repeatability results for different combinations of handcrafted filters and a number of learnable layers ($M=8$ filters each). A higher number of layers leads to better results. All repeatability scores are computed on synthetic validation set from ImageNet.}
\label{table:context_losses_and_learnableblocks}
\end{figure*}


\subsection{Implementation Notes}
\label{subsec:Implementation_Details}
Training is performed in a siamese pipeline, with two instances of Key.Net that share the weights and are updated at the same time.
Each convolutional layer has $M$ = 8 filters of size $5 \times 5$, with He weights initialization and L2 kernel regularizer. 
We compute the covariant constraint loss $\mathcal{L}_{M\mbox{-}SIP}$ for five scale levels, with the size of the M-SIP windows $N_s \in [8, 16, 24, 32, 40]$ and loss term $\lambda_s \in [256, 64, 16, 4, 1]$, that were determined by performing a hyperparameter search on the validation set. Larger candidate window sizes have greater mean errors between coordinate points since the maximum distance is proportional to the window size. Thus, $\lambda_s$ has the largest value for the smallest window. We use a batch size of 32, an Adam Optimizer with a learning rate of $10^{-3}$ and a decay factor of 0.5 after 20 epochs. On average, the architecture converges in 30 epochs, 2h on a machine with an i7-7700 CPU running at 3.60GHz and a NVIDIA GeForce GTX 1080 Ti. Evaluation benchmark, synthetic data generator, Key.Net network, and loss are implemented using TensorFlow and are available on GitHub\footnote{https://github.com/axelBarroso/Key.Net}.

\begin{table}[h]
    \begin{subtable}[h]{0.45\textwidth}
    \begin{center}
    \begin{tabular}{lcccccc}
    \hline
    \noalign{\smallskip}
    \multicolumn{1}{c}{} & \multicolumn{6}{c}{Num. Pyramid Levels} \\ 
    \cline{2-7} \noalign{\smallskip}
    & 1 & 2 & 3 & 4 & 5 & 6\\
    \hline
    \noalign{\smallskip}
    Rep. & 72.5 & 74.6 & 79.1 & 79.4 & \textbf{79.5} & 78.6  \\
    \vspace{-0.9cm}
    \end{tabular}
    \end{center}
    \caption{Number of input scale levels in Key.Net.}
    \label{table:pyramid_levels}
    \end{subtable}
    \hfill
    \vspace{0.1cm}
    \begin{subtable}[h]{0.45\textwidth}
    \begin{center}
    \begin{tabular}{lcccccc}
    \hline
    \noalign{\smallskip}
        \multicolumn{1}{c}{} & \multicolumn{6}{c}{Spatial Softmax Base} \\ 
        \cline{2-7} \noalign{\smallskip}
        & 1.2 & 1.4 & 2.0 & $e$ & 5.0 & 7.5 \\
        \hline
        \noalign{\smallskip}
        Rep. & 77.5 & 78.4 & 77.9 & \textbf{79.1} & 74.6 & 73.2 \\
        \vspace{-0.9cm}
        \end{tabular}
        \end{center}
        \caption{Spatial softmax base used in equation~\ref{eq:spatial_softmax}.}
        \label{table:spatial_softmax}
     \end{subtable}
     \vspace{-0.2cm}
     \caption{Repeatability results for different design choices on synthetic validation set from ImageNet.}
     \label{tab:design_choices}
\end{table}

\section{Results}
\label{sec:results}

In this section, we present the experiments and discuss the results. We first show results on validation data for several variants of the proposed architecture. Next, Key.Net repeatability scores in single-scale and multi-scale are presented along with the state-of-the-art detectors on HPatches. Moreover, we evaluate the matching performance, the number of learnable parameters and inference time of our proposed detector and compare to other techniques.


\subsection{Preliminary Analysis}

\begin{table*}[th]
\vspace{-0.10cm}
\begin{center}
\begin{tabular}{lccccccclccccccc}
\hline
\noalign{\smallskip}
\multicolumn{1}{c}{} & \multicolumn{7}{c}{Viewpoint} & \multicolumn{1}{c}{} & \multicolumn{7}{c}{Illumination} \\ 
\cline{2-8} \cline{10-16} \noalign{\smallskip}
\multicolumn{1}{c}{} & \multicolumn{2}{c}{Repeatability} && \multicolumn{2}{c}{$\Bar{\epsilon}_{IoU}$} && \multicolumn{1}{c}{$S_{range}$} & \multicolumn{1}{c}{} & \multicolumn{2}{c}{Repeatability} && \multicolumn{2}{c}{$\Bar{\epsilon}_{IoU}$} && \multicolumn{1}{c}{$S_{range}$} \\
\cline{2-3} \cline{5-6} \cline{8-8} \cline{10-11}  \cline{13-14} \cline{16-16} \noalign{\smallskip}
 & SL & L && SL & L && SL && SL & L && SL & L && SL \\
\noalign{\smallskip}
\hline
\noalign{\smallskip}
SIFT-SI \cite{DoG}                    & 43.1 & 57.6 && \textbf{\textcolor{darkgray}{\underline{0.18}}} & 0.12 && 78.6 && 47.8 & 60.4 && 0.18 & 0.12 && 84.5 \\
SURF-SI \cite{SURF}                    & 46.7 & 60.3 && \textbf{\textcolor{darkgray}{\underline{0.18}}} & 0.18 && 24.8 && 53.0 & 64.0 && 0.15 & 0.11 && 27.4 \\
FAST-TI \cite{FAST}                    & 30.4 & 63.1 && 0.21 & \textbf{\textcolor{darkgray}{\underline{0.10}}} && - && 63.6 & 63.6 && \textbf{0.09} & \textbf{\textcolor{darkgray}{\underline{0.09}}} && - \\
MSER-SI \cite{MSER}                    & 56.4 & 62.8 && \textbf{0.12} & \textbf{0.08} && \textbf{503.7} && 46.5 & 54.5 && 0.12 & 0.10 && \textbf{524.8} \\
Harris-Laplace-SI \cite{HarrisLaplace} & 45.1 & 62.0 && 0.20 & 0.13 && \textbf{\textcolor{darkgray}{\underline{95.9}}} && 52.7 & 62.0 && 0.17 & \textbf{0.08} && \textbf{\textcolor{darkgray}{\underline{90.4}}} \\
KAZE-SI \cite{KAZE}                    & 53.3 & 65.7 && 0.20 & 0.11 && 12.5 && 56.9 & 65.7 && 0.12 & 0.10 && 12.7 \\
AKAZE-SI \cite{AKAZE}                  & 54.0 & 65.6 && 0.19 & \textbf{\textcolor{darkgray}{\underline{0.10}}} && 13.5 && 64.9 & 69.1 && 0.11 & \textbf{\textcolor{darkgray}{\underline{0.09}}} && 13.6 \\
TILDE-TI \cite{TILDE}                  & 31.0 & 65.1 && 0.20 & 0.15 && - && \textbf{\textcolor{darkgray}{\underline{70.4}}} & 70.4 && 0.11 & 0.11 && - \\
LIFT-SI \cite{LIFT}                    & 43.4 & 59.4 && 0.20 & 0.13 && 13.3 && 51.6 & 65.4 && 0.18 & 0.12 && 13.8 \\
DNet-SI \cite{Karel_Vedaldi_ECCV_16}   & 49.4 & 62.2 && 0.21 & 0.14 && 11.4 && 59.1 & 65.1 && 0.14 & 0.13 && 17.1 \\
TCDET-SI \cite{Zhang_Felix_CVPR_17}    & 49.6 & 61.6 && 0.23 & 0.16 && 6.7 && 66.9 & \textbf{\textcolor{darkgray}{\underline{71.0}}} && 0.16 & 0.15 && 11.4 \\
SuperPoint-TI \cite{detone2017superpoint} & 33.3 & 67.1 && 0.20 & 0.17 && - && 69.9 & 69.9 && \textbf{\textcolor{darkgray}{\underline{0.10}}} & 0.10 && - \\
LF-Net-SI \cite{OnoSerra18}            & 32.3 & 62.2 && 0.23 & 0.12 && 2.00 && 68.6 & 69.1 && \textbf{\textcolor{darkgray}{\underline{0.10}}} & 0.10 && 2.0 \\
\noalign{\smallskip}
\hline
\noalign{\smallskip}
Tiny-Key.Net-SI & \textbf{\textcolor{darkgray}{\underline{57.8}}} & 70.3 && 0.20 & 0.12 && 7.6 && 56.1 & 62.8 && 0.14 & 0.11 && 7.6 \\
Key.Net-TI & 34.2 & \textbf{\textcolor{darkgray}{\underline{71.5}}} && 0.20 & 0.11 && - && \textbf{72.0} & \textbf{72.0} && \textbf{\textcolor{darkgray}{\underline{0.10}}} & 0.10 && - \\


Key.Net-SI & \textbf{60.5} & \textbf{73.2} && 0.19 & 0.14 && 7.6 && 61.3 & 66.2 && 0.12 & 0.10 && 7.6 \\
\end{tabular}
\vspace{-0.2cm}
\caption{
Repeatability results (\%) for translation (TI) and scale (SI) invariant detectors on HPatches. We also report average overlap error $\Bar{\epsilon}_{IoU}$ and ratio of maximum to minimum extracted scale $S_{Range}$. In SL, scales and locations are used to compute overlap error, meanwhile, in L, only locations are used and scales are assumed to be correctly estimated. Key.Net and Tiny-Key.Net are the best algorithms on viewpoint, for both L and SL. On illumination sequences, translation invariant Key.Net-TI obtains the best accuracy. Among scale invariant SI detectors, TCDET is the best in L and LF-Net in SL.}
\label{table:repeatability_scores}
\end{center}
\end{table*}

\label{subsec:preliminary_analysis}
We study several combinations of loss terms, different handcrafted filters and the effects of the number of learnable layers or pyramid levels within the architecture. \par
\noindent\textbf{M-SIP Levels} are investigated in figure~\ref{table:context_losses_and_learnableblocks} (Left) showing increasing repeatability with more scale levels within M-SIP operator. In addition, we show how the loss with smaller window size $N$ improves repeatability. However, the best result is obtained when all levels are combined. 

\noindent\textbf{Filter Combinations} are analyzed in figure~\ref{table:context_losses_and_learnableblocks} (Right).
We show results for $1^{st}$ and $2^{nd}$ order filters as well as their combination. All networks have the same number of filters, however, we either freeze first layer of 10 filters with handcrafted kernels (c.f. section \ref{sec:handcrafted_filters}) or learn them depending on the variant of our network, e.g, in Fully Learnable Key.Net there are no handcrafted filters as all are randomly initialized and learned.  
The results show that the information provided by handcrafted filters is essential when the number of learnable layers is small. Handcrafted filters act as soft constraints, which directly discard areas without gradients, i.e. non-discriminative with low repeatability. However, as we add more learnable blocks, repeatability scores for combined and fully learnable networks become comparable. Naturally, gradient-based handcrafted filters are simple, and architectures with enough complexity could learn them if they were required. However, the use of engineered features leads to a smaller architecture while maintaining the performance, which is often critical for real-time applications. In summary, combining both types of filters allows to significantly reduce the number of learnable layers. We use Key.Net architecture with three learnable blocks in the next experiments. \par

\noindent\textbf{Multiple Pyramid Levels} at the input to the network also affect the detection performance as shown in table~\ref{table:pyramid_levels}. For a single pyramid level, only the original image is used as input. Adding pyramid levels is similar to increasing the size of the receptive fields in the architecture. Our experiment suggests that using more than three levels does not lead to significantly improved results. On the validation set, we obtain a repeatability score of 72.5\% for one level,  an increase of 6.6\% for three, and 7.0\% for five levels. We, therefore, use three levels, which achieve good performance while keeping the computational cost low. \par
\noindent\textbf{Spatial Softmax Base} in equation~\ref{eq:spatial_softmax} defines how {\em soft} the estimation of keypoint coordinates  is. High values return the location of the global maximum within the window, while low values average local maxima. The base is varied in table~\ref{table:spatial_softmax}. 
Optimum scores are obtained when using the base in equation~\ref{eq:spatial_softmax}  close to the $e$ value, which is in line with the setting used in \cite{keypointnet_nips_18}.

\subsection{Keypoint Detection}

This section presents the results for state-of-the-art local feature detectors along with our proposed method. Table \ref{table:repeatability_scores} shows the repeatability score,  average intersection-over-union error $\Bar{\epsilon}_{IoU}$ and scale range $S_{range}$, which is the ratio between the maximum and minimum scale values of the extracted interest points.  Suffixes -TI and -SI, refer to translation (detection at a single scale only) and scale invariance (detection at multiple scales), respectively. Keypoint location is only evaluated under L by assuming correct scale detection, while scale and location (SL) use the actual detected scale and location for computing the repeatability and overlap error. \par
In addition to Key.Net, we propose Tiny-Key.Net, which is a reduced size architecture with all handcrafted filters but only one learnable layer with one filter  ($M = 1$) and a single scale input. The idea behind Tiny-Key.Net is to demonstrate how far the complexity can be reduced while keeping good performance. Key.Net and Tiny-Key.Net are extended to scale invariance by evaluating the detector on several scaled images, similar to \cite{Zhang_Felix_CVPR_17}. We also show results on single scale input Key.Net-TI, to compare it directly with other TI detectors such as SuperPoint or TILDE. We set the thresholds of algorithms such that they return at least 1,000 points per image. As MSER proposes regions without scoring or ranking, we randomly pick 1,000 points to compute the results. We repeat this experiment ten times and average the results for MSER. Key.Net has the best results on viewpoint sequences, in terms of both, location and scale. Tiny-Key.Net does not perform as well as Key.Net but it is within the top three repeatability scores, after Key.Net-TI and Key.Net-SI. \newline
On illumination sequences, Key.Net-TI performs the best among TI detectors, not being affected by scale estimation errors. TCDET, which uses points detected by TILDE as anchors, is the most accurate in location estimation compared to other SI detectors. Note that TILDE based detectors were specifically designed and trained for illumination sequences. LF-Net is the best SI detector according to SL overlap, not suffering much from incorrect scale estimations. However, its repeatability decreases the most from L to SL among all SI detectors on viewpoint sequences. Key.Net-SI addresses the scale changes better than the other methods but the errors in multi-scale sampling affect it when there is no scale change between images i.e. illumination sequences. This has often been observed for detectors with more invariance than required by the data. Handcrafted detectors have the lowest average overlap error $\Bar{\epsilon}_{IoU}$ among all detectors. A wide range of scales $S_{range}$ is detected by MSER, which has a great capability of extracting local features from different scales due to its feature segmentation nature.

\begin{table}
\vspace{-0.10cm}
\begin{center}
\begin{tabular}{lcc}
\hline
\noalign{\smallskip}
\multicolumn{1}{c}{} & \multicolumn{2}{c}{Matching Score} \\ 
\cline{2-3} \noalign{\smallskip}
\multicolumn{1}{c}{} & \multicolumn{1}{c}{View} & \multicolumn{1}{c}{Illum} \\
\noalign{\smallskip}
\hline
\noalign{\smallskip}
MSER \cite{MSER} + HardNet \cite{Mishchuk_Mishkin_NIPS17}   & 11.7 & 18.8 \\
SIFT \cite{DoG} + HardNet \cite{Mishchuk_Mishkin_NIPS17}   & 23.2 & 24.8 \\
HarrisLaplace \cite{HarrisLaplace} + HardNet \cite{Mishchuk_Mishkin_NIPS17} & 30.0 & 31.7 \\
AKAZE \cite{AKAZE} + HardNet \cite{Mishchuk_Mishkin_NIPS17} & 36.4 & 41.4 \\
TILDE \cite{TILDE} + HardNet \cite{Mishchuk_Mishkin_NIPS17} & 32.3 & 39.3 \\
LIFT \cite{LIFT} + HardNet \cite{Mishchuk_Mishkin_NIPS17} & 30.3 & 32.8 \\
DNet \cite{Karel_Vedaldi_ECCV_16} + HardNet \cite{Mishchuk_Mishkin_NIPS17} & 33.5 & 34.7 \\
TCDET \cite{Zhang_Felix_CVPR_17} + HardNet \cite{Mishchuk_Mishkin_NIPS17} & 27.6 & 36.3 \\
SuperPoint \cite{detone2017superpoint} + HardNet \cite{Mishchuk_Mishkin_NIPS17} & 37.4 & \textbf{\textcolor{darkgray}{\underline{43.0}}} \\
LF-Net \cite{OnoSerra18} + HardNet \cite{Mishchuk_Mishkin_NIPS17} & 26.9 & \textbf{43.8} \\
\noalign{\smallskip}
\hline
\noalign{\smallskip}
LIFT \cite{LIFT} & 21.8 & 26.5 \\
SuperPoint \cite{detone2017superpoint} & \textbf{\textcolor{darkgray}{\underline{38.0}}} & 41.5 \\
LF-Net \cite{OnoSerra18} & 23.0 & 29.1 \\
\noalign{\smallskip}
\hline
\noalign{\smallskip}
Tiny-Key.Net + HardNet \cite{Mishchuk_Mishkin_NIPS17} & 37.9 & 37.3 \\
Key.Net + HardNet \cite{Mishchuk_Mishkin_NIPS17}      & \textbf{38.4} & 39.7 \\
\end{tabular}
\vspace{-0.2cm}
\caption{Matching score (\%) of best detectors together with HardNet and state-of-the-art detector/descriptors. Results on HPatches sequences, both viewpoint, and illumination. Key.Net architecture gets the best matching score for viewpoint, while LF-Net+HardNet for illumination sequences.}
\label{table:matchingScore}
\end{center}
\end{table}

\subsection{Keypoint Matching}
Moreover, in order to demonstrate that the detected features are useful for matching, table \ref{table:matchingScore} shows matching scores for detectors combined with HardNet  descriptor \cite{Mishchuk_Mishkin_NIPS17}. As our method only focuses on the detection part, and for a fair comparison, we used the same descriptor and discard the orientation for all methods that provide it. In addition, we include in the table LIFT \cite{LIFT}, SuperPoint \cite{detone2017superpoint} and \mbox{LF-Net \cite{OnoSerra18}} with their descriptors, but ignoring their orientation estimation. SuperPoint and LF-Net have 256 descriptor dimension, while dimension of HardNet \cite{Mishchuk_Mishkin_NIPS17} and LIFT is 128. Matching score is computed as the ratio between features matched and detected (top 1,000). Top matching scores is obtained by Key.Net on viewpoint, and \mbox{LF-Net+HardNet} on illumination. Feature detectors that were optimized jointly with a descriptor \cite{LIFT, detone2017superpoint, OnoSerra18} have better matching score than regular learned detectors on illumination sequences, but not on viewpoint. Handcrafted AKAZE performs close to the top learned methods for both viewpoint and illumination sequences.\par

\subsection{Efficiency}
We also compare the number of learnable parameters, indicating then the complexity of the predictor, which leads to an increasing risk of overfitting and need for a large amount of training data. Table~\ref{table:number_parameters} shows the approximate number of parameters for different architectures. 
Learnable parameters that are not used during inference in the detector part are not counted for SuperPoint and LF-Net detectors.
The highest complexity is from SuperPoint with $940k$ learnable parameters. Key.Net has nearly 160 times fewer parameters and Tiny-Key.Net has $3$,$100$ times fewer parameters than SuperPoint with better repeatability for viewpoint scenes.
The inference time of an image of 600 $\times$ 600 is 5.7ms (175 FPS) and 31ms (32.25 FPS) for Tiny-Key.Net and Key.Net, respectively.

\begin{table}[t]
\vspace{-0.10cm}
\begin{center}
\begin{tabularx}{\linewidth}{ccccc}
\hline
\noalign{\smallskip}
\multicolumn{5}{c}{Number of Learnable Parameters} \\ 
\cline{1-5} \noalign{\smallskip}
\scalebox{0.92}{TCDET} & \scalebox{0.92}{SuperPoint} & \scalebox{0.92}{LF-Net} & \scalebox{0.92}{Tiny-Key.Net} & \scalebox{0.92}{Key.Net}\\
\hline
\noalign{\smallskip}
548k & 940k & 39k & \textbf{280} & \textbf{\textcolor{darkgray}{\underline{5.9k}}} \\
\noalign{\smallskip}
\noalign{\smallskip}
\vspace{-0.75cm}
\end{tabularx}
\vspace{-0.2cm}
\caption{Comparison of the number of learnable parameters for state-of-the-art architectures. Tiny-Key.Net has only one learnable block with one filter.}
\label{table:number_parameters}
\end{center}
\end{table}

\section{Conclusions}
\label{sec:Conclusions}

We have introduced a novel approach to detect local features that combines handcrafted and learned CNN filters.  We have proposed a multi-scale index proposal layer that finds keypoints across a range of scales, with a loss function that optimizes the robustness and discriminating properties of the detections. We demonstrated how to compute and combine differentiable keypoint detection loss for multi-scale representation. Evaluation results on large benchmark show that combining handcrafted and learned features as well as multi-scale analysis at different stages of the network improves the repeatability scores compared to other state-of-the-art keypoint detection methods.

We further show that excessively increasing network's complexity does not lead to improved results. In contrast, using handcrafted filters allows to significantly reduce the complexity of the architecture leading to a detector with 280 learnable parameters and inference of 175 frames per second. Proposed detectors lead to state-of-the-art matching performance when used with a descriptor on viewpoint. 

{\small
\bibliographystyle{unsrt}
\bibliographystyle{ieee_fullname}
\bibliography{egbib}
}

\end{document}